\documentclass[11pt]{article}

\usepackage[preprint]{acl}

\usepackage{times}
\usepackage{latexsym}
\usepackage{booktabs}       
\usepackage[T1]{fontenc}

\usepackage[utf8]{inputenc}
\usepackage{multirow}
\usepackage{microtype}
\usepackage[ruled,vlined,linesnumbered]{algorithm2e}
\usepackage{inconsolata}
\usepackage{amsmath}
\usepackage{graphicx}

\title{DiffuMask: Diffusion Language Model for Token-level Prompt Pruning}

\author{
 \textbf{Caleb Zheng\textsuperscript{1}},
 \textbf{Jyotika Singh\textsuperscript{2}},
 \textbf{Fang Tu\textsuperscript{2}},
 \textbf{Weiyi Sun\textsuperscript{2}},
\\
 \textbf{Sujeeth Bharadwaj\textsuperscript{2}},
 \textbf{Yassine Benajiba\textsuperscript{2}},
 \textbf{Sujith Ravi\textsuperscript{2}},
\\
\textbf{Eli Shlizerman \textsuperscript{1}},
 \textbf{Dan Roth\textsuperscript{2}},
\\
\\
 \textsuperscript{1} University of Washington,
 \textsuperscript{2}Oracle AI
\\
 \small{
   \textbf{Correspondence:} \href{zheng94@uw.edu}{zheng94@uw.edu}
 }
}

\begin{document}
\maketitle
\begin{abstract}
In-Context Learning and Chain-of-Thought prompting improve reasoning in large language models (LLMs). These typically come at the cost of longer, more expensive prompts that may contain redundant information. Prompt compression based on pruning offers a practical solution, yet existing methods rely on sequential token removal which is computationally intensive. We present \textsc{DiffuMask}, a diffusion-based framework integrating hierarchical shot-level and token-level pruning signals, that enables rapid and parallel prompt pruning via iterative mask prediction. \textsc{DiffuMask} substantially accelerates the compression process via masking multiple tokens in each denoising step. It offers tunable control over retained content, preserving essential reasoning context and achieving up to 80\% prompt length reduction. Meanwhile, it maintains or improves accuracy across in-domain, out-of-domain, and cross-model settings. Our results show that \textsc{DiffuMask} provides a generalizable and controllable framework for prompt compression, facilitating faster and more reliable in-context reasoning in LLMs.
\end{abstract}

\section{Introduction}
Large language models (LLMs) excel at tasks such as text generation, translation, and complex reasoning. A key capability underlying this success is In-Context Learning (ICL), which allows models to adapt to novel tasks using only natural language demonstrations without parameter updates~\cite{brown2020language}. Furthermore, the Chain-of-Thought (CoT) paradigm~\cite{wei2022emergent} has been shown empirically to improve ICL performance in many settings by encouraging models to generate intermediate textual steps. While the precise mechanisms underlying these gains remain an open question, extensive empirical evidence shows that prompting models to generate explicit reasoning traces can significantly improve performance across a wide range of tasks~\cite{wei2022emergent, kojima2022large, wang2022self, zhou2023leasttomost}. Meanwhile, these benefits are not universal: additional context does not always translate to better performance, and redundant steps or irrelevant exemplars can obscure useful signals and even degrade accuracy~\cite{liu2024lost}. Moreover, longer prompts inevitably increase inference latency and memory usage, highlighting a fundamental trade-off between \textit{performance gains} from richer context and the \textit{efficiency cost} of prompt length.

Prompt compression has thus emerged as a promising solution to reduce prompt length without compromising model performance. We focus on compressing the dynamic run-time portion of the prompt, such as task instructions, exemplars, and reasoning traces, after prompt construction and prior to inference, where prompt length most directly affects efficiency. Existing approaches fall into two main categories: (i) soft compression, which creates dense embeddings that summarize the prompt’s semantics~\cite{wingate2022prompt}, and (ii) hard compression, which directly manipulates discrete tokens to remove redundancies~\cite{li2023compressing, li2024prompt, jung2024discrete, shandilya2025taco}. Hard compression methods can be divided into summarization-based and pruning-based approaches. The former rewrites the prompt into a shorter textual form, which can sometimes compromise semantic fidelity, while the latter explicitly removes redundant tokens, preserving most of the original linguistic structure and interpretability. While effective, existing pruning-based methods rely on sequential or reinforcement-driven mechanisms that are slow and difficult to scale.

To address these limitations, we propose \textsc{DiffuMask}, a diffusion-based framework for efficient, scalable, and controllable prompt compression. \textsc{DiffuMask} frames prompt pruning as an iterative denoising process, predicting token retention masks in parallel and dramatically reducing computational overhead while preserving essential reasoning context,  offering a new perspective on prompt optimization.

Our main contributions are as follows:
\begin{itemize}
\item We introduce a \textit{diffusion-based} framework for prompt pruning that provides an efficient and scalable alternative to sequential methods.\vspace{-0.2cm}
\item We present the pipeline of constructing a \textit{full-pruned prompt dataset} combining hierarchical shot- and token-level pruning, demonstrating its application on the GSM8K mathematical reasoning benchmark~\cite{cobbe2021training}.\vspace{-0.2cm}
\item We perform a detailed empirical evaluation on GSM8K, demonstrating that \textsc{DiffuMask} achieves up to \textit{$80\%$} prompt length reduction within $1$ minute. This provides acceleration of orders-of-magnitude over sequential pruning methods which typically require 10--48 hours, while maintaining or improving task accuracy.\vspace{-0.2cm}
\item We demonstrate the generalization ability of \textsc{DiffuMask} through \textit{out-of-domain and cross-model evaluations}, showing that it transfers effectively across tasks and architectures, enabling broader adaptability in practical LLM applications.
\end{itemize}

\section{Related Work}

Recent advances in LLMs have driven substantial research on improvement of reasoning efficiency and on prompt design. Here, we review three closely related directions: (1) In-Context Learning (ICL) and Chain-of-Thought (CoT) reasoning, (2) Prompt Compression, and (3) Diffusion Language Models (DLMs).

\paragraph{ICL and CoT Reasoning.}
LLMs such as GPT-3~\cite{brown2020language} first demonstrated ICL, enabling models to adapt to new tasks from just a few demonstrations without finetuning. Subsequent work has explored how demonstration selection and ordering affect ICL performance~\cite{liu2021makes, sorensen2022information}.
While ICL facilitates adaptation at inference time, it does not inherently support multi-step reasoning. The CoT prompting paradigm~\cite{wei2022emergent, wei2022chain} addresses this by guiding models to produce intermediate reasoning steps, substantially boosting performance on complex tasks. CoT can be applied in both few-shot and zero-shot settings~\cite{kojima2022large}, and recent theoretical work suggests that these intermediate steps enhance the computational expressiveness of transformer models~\cite{merrill2024the, yuan2026barriersdiscretereasoningtransformers, singh2023}.

Despite these advances, LLMs exhibit strong sensitivity to prompt structure and token ordering~\cite{s-etal-2025-llms}. Multiple observations show that misplacement of salient information can substantially degrade performance~\cite{liu2024lost}. As few-shot CoT prompts grow longer and thus more prone to content redundancy, development of efficient prompt optimization methods becomes increasingly critical to balance prompt length with performance. Prompt compression, in particular, provides a promising avenue to address this challenge.
\paragraph{Prompt Compression.}
Prompt compression methods fall into two types: soft and hard compression. Soft compression learns compact embeddings that replace explicit tokens~\cite{wingate2022prompt, gecontext, mu2023learning}. In contrast, hard compression works directly with discrete tokens, producing concise and interpretable prompts~\cite{chen2023walking, yoon2024compact}.

Hard compression based on pruning removes redundant tokens~\cite{li2023compressing, jiang2023llmlingua, pan2024llmlingua}. Notable examples include \textsc{CoT-Influx}~\cite{huang2024fewer}, which employs hierarchical shot-level and token-level pruning, and \textsc{PromptQuine}~\cite{wang2025evolving}, which formulates prompt optimization as an open-ended evolutionary search. While these approaches achieve impressive compression ratios and maintain task accuracy, they often require task-specific adaptation and rely on sequential procedures which is computationally intensive. Such procedure can take from several hours to days to process a single prompt which limits their scalability to diverse tasks or large datasets, underscoring the need for a more efficient and scalable pruning framework.

\paragraph{DLMs.}
Diffusion models have achieved remarkable success in image generation~\cite{ho2020denoising, song2021scorebased} and have recently been extended to language modeling. Unlike autoregressive (AR) models that generate text sequentially, DLMs perform parallel denoising across all tokens, enabling faster inference, bidirectional context utilization, and robustness to exposure bias.
\begin{figure*}
    \centering
    \includegraphics[width=1.0\linewidth]{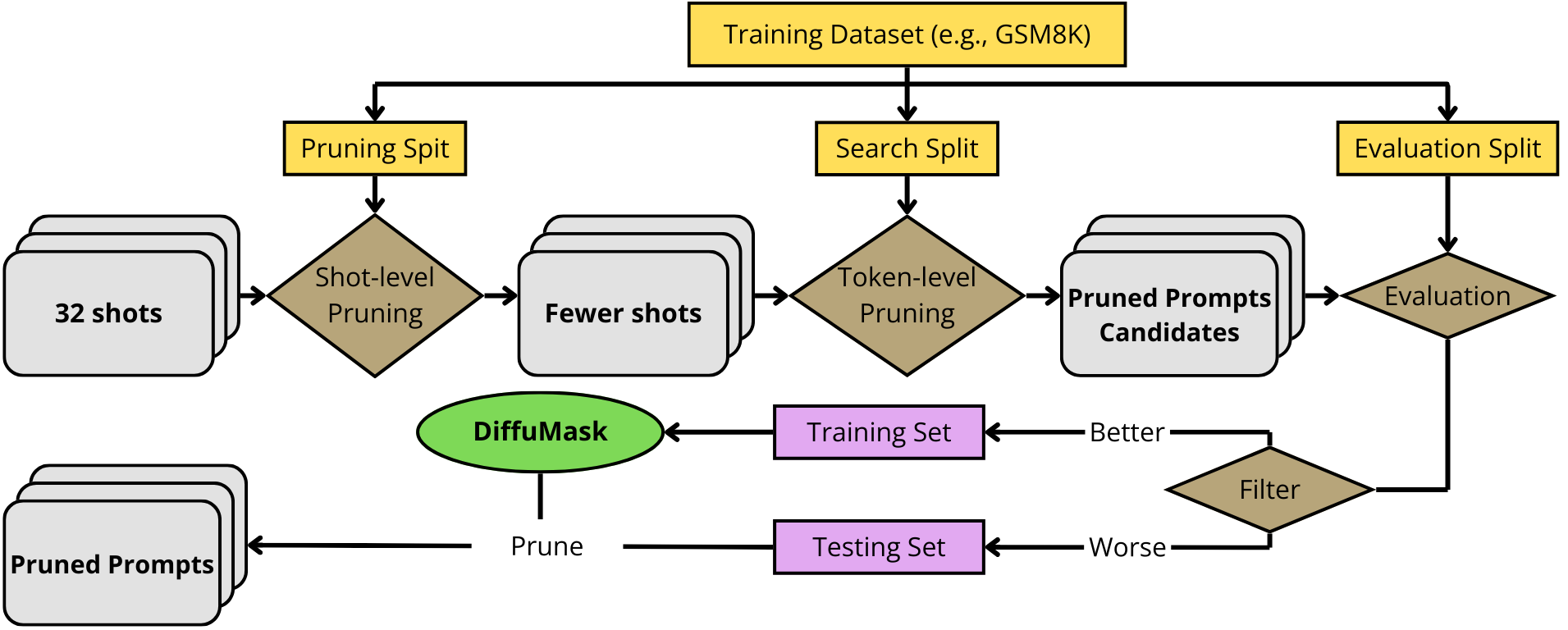}
    \caption{Data generation pipeline for \textsc{DiffuMask}. The method applies hierarchical shot-level and token-level pruning to generate pruned prompt candidates, which are evaluated and filtered into training and validation sets. GSM8K is shown as a representative example; the pipeline is applicable to any dataset with long prompts.}
    \label{fig:pipeline}
\end{figure*}

\textsc{Diffusion-LM}~\cite{li2022diffusion} pioneered the continuous DLM paradigm which corrupts token embedding with Gaussian noise and trains models to reconstruct clean sequences. Subsequent work ~\cite{gong2023diffuseq, dieleman2022continuous, lin2023text} explored improvements in architecture and training objectives, yet these continuous DLMs faced scalability challenges on large text corpora. Discrete DLMs were proposed, applying discrete corruption processes directly at the token level to enable better alignment with linguistic structure and enhance scalability~\cite{austin2021structured}. Further advancements have improved the training stability and performance of discrete DLMs~\cite{lou2023discrete, ou2025your, gongscaling, nie2025scaling}, narrowing the gap between diffusion- and AR-based language models. Recent large-scale discrete DLMs—including \textsc{Dream-7B}~\cite{ye2025dream} and \textsc{LLaDA-8B}~\cite{nie2025large}—demonstrate that diffusion-based architectures can scale to contemporary LLM sizes and match or surpass AR baselines on various tasks.

Beyond scalability, DLMs show pronounced data efficiency: they benefit from repeated exposure to limited training data and outperform AR baselines in data-constrained regimes~\cite{prabhudesai2025diffusion}. This property is especially relevant for prompt compression, where large-scale full-prompt $\rightarrow$ pruned-prompt pairs are scarce and conventional compression methods are expensive to deploy.

These findings position DLMs as a scalable, data-efficient alternative for prompt pruning, forming the methodological foundation for our approach.
\section{Methods}

We introduce \textsc{DiffuMask}, a finetuned diffusion language model (DLM) that predicts token retention masks in parallel, enabling efficient and scalable prompt compression. Its effectiveness depends on high-quality supervision that captures realistic hierarchical pruning behaviors and provides reliable mask prediction. This section first describes the construction of the full–pruned prompt dataset used to train the \textsc{DiffuMask} model, followed by a detailed explanation of the framework and its differences from conventional masked DLMs. The overall procedure is illustrated in Figure~\ref{fig:pipeline}.

\subsection{Dataset Construction}
\label{section:dataset}
The dataset construction process is based on the GSM8K math reasoning benchmark, commonly used for ICL and CoT reasoning. To model realistic hierarchical pruning, the process involves \emph{shot-level pruning}, which removes entire exemplars, and \emph{token-level pruning}, which eliminates redundant tokens within them. Specifically, our approach is informed by two representative prompt optimization methods: \textsc{CoT-Influx}~\cite{huang2024fewer}, which trains a shot-level pruner using reinforcement learning (RL), and \textsc{PromptQuine}~\cite{wang2025evolving}, which performs token-level pruning through evolutionary search.

Our dataset construction pipeline proceeds through four sequential stages. First, we perform \emph{shot-level pruning} to reduce the number of in-context exemplars using both trained and random selection strategies. Next, \emph{token-level pruning} refines these prompts to achieve finer-grained compression. The third stage, \emph{evaluation and filtering}, identifies high-quality pruned prompts that improve accuracy compared to their full counterparts. Finally, \emph{data representation and alignment} formalizes the dataset into a consistent supervision format for DiffuMask training. More details about each stage implementation is described in Appendix.~\ref{appendix:dataset_construction}.

To support these stages, we divide the GSM8K dataset into three subsets: a \emph{pruning split} (6{,}000 samples) for training the shot-level pruner, a \emph{search split} (200 samples) for evolutionary search, and an \emph{evaluation split} (1{,}273 samples) for assessing prompt quality and selecting final pruned prompts. 

\paragraph{Shot-level Pruning} 
We use the pruning split to train the shot-level pruner. Following the setup of CoT-Influx, we divide the pruning split into a candidate set and a query set. In each RL training episode, one query is randomly selected from the query set and retrieves 32 candidate exemplars from the candidate set based on their semantic similarity to the query. The pruner is trained for three epochs to optimize its selection policy.
After training, we select 100 32-shot prompts as full prompts and apply the trained pruner to them, retaining an average of approximately 10 shots per prompt. We refer to these shot-level pruned prompts as \emph{fewer-shot prompts}, which achieve comparable or slightly higher accuracy than the full prompts while substantially reducing prompt length. 

It is worth noting that CoT-Influx serves as just one possible approach for the shot-level pruning stage; other selection methods can also be used. To explore this flexibility, we employ two types of random selection. The first variant randomly selects a variable number of $k$ shots from the full prompt, where $k$ differs across prompts but is sampled such that the average number of remaining shots matches that of the trained pruner. This produces 100 additional prompt pairs. The second variant randomly samples 32 shots from the pruning split, then selects a fixed number of $k$ shots from these to simulate randomly composed prompts. We evaluate multiple values of $k$ and find that $k=5$ achieves the best overall performance. Consequently, we adopt $5$-shot pruning for $600$ additional prompts, resulting in a total of 800 fewer-shot prompts in our dataset.

\paragraph{Token-level Pruning}
We then perform \emph{token-level pruning} on those fewer-shot prompts to achieve finer-grained compression while preserving essential reasoning content. We adopt \textsc{TA-Pruning}, a simplified variant derived from the PromptQuine framework. While PromptQuine combines multiple mechanisms-including self-replication, mutation, and prompt ranking-to iteratively evolve prompts, \textsc{TA-Pruning} applies only the threshold-accepting component. Specifically, it iteratively removes tokens whose deletion does not degrade task accuracy beyond a fixed tolerance. Despite its much lower complexity, \textsc{TA-Pruning} achieves comparable pruning performance with minimal hyperparameter sensitivity.

The \emph{search split} is used to evaluate candidate pruning steps and guide the pruning trajectory. In this setting, \textsc{TA-Pruning} significantly reduces prompt length while maintaining—or even improving— accuracy. However, its sequential nature makes the algorithm computationally expensive and difficult to parallelize. In practice, processing a single fewer-shot prompt can take between 10 hours and 2 days on an A100 GPU, depending on prompt length and pruning trajectory. This computational overhead renders TA-Pruning impractical for large-scale prompt optimization or inference-time pruning, motivating our development of \textsc{DiffuMask} as an efficient, scalable alternative.
\begin{figure*}
    \centering
    \includegraphics[width=1.0\linewidth]{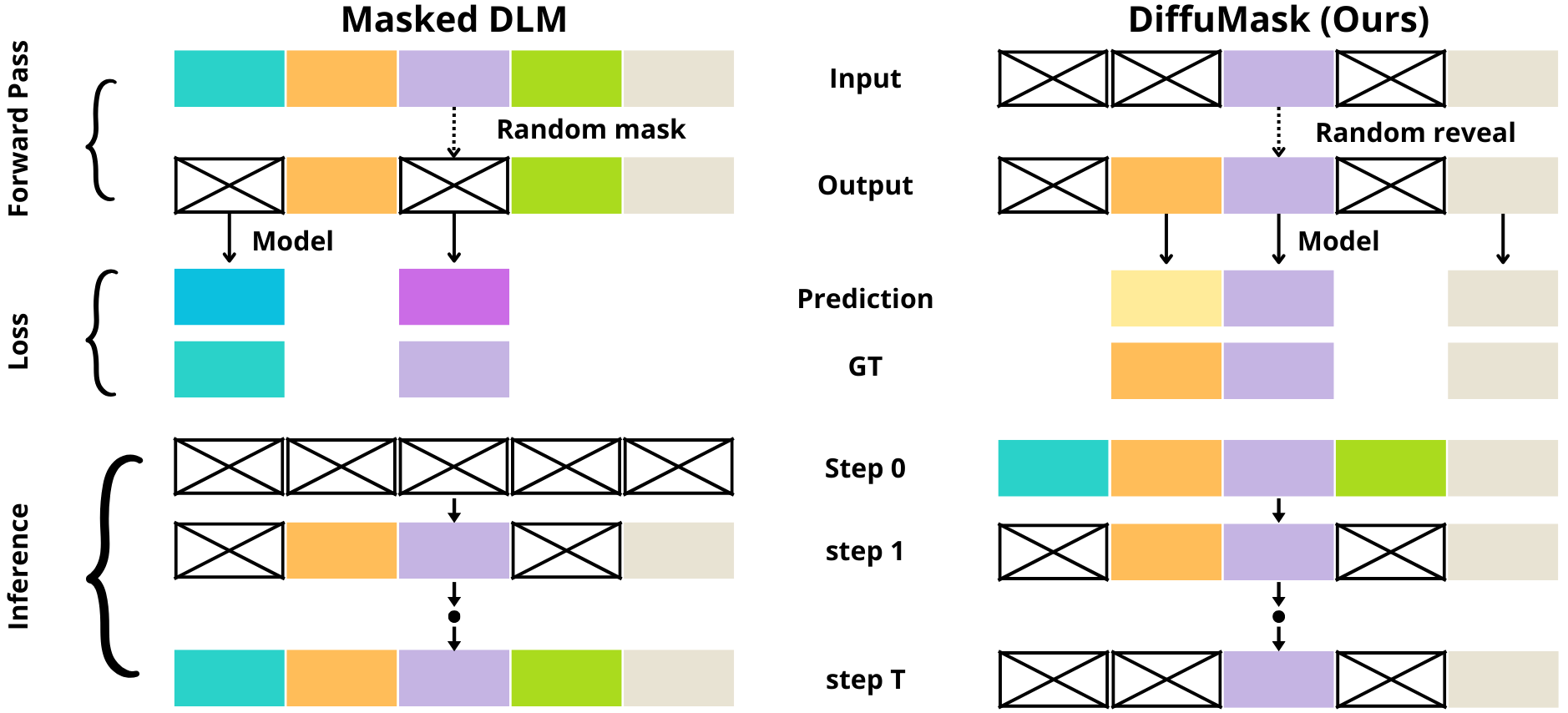}
    \caption{Overview of the proposed DiffuMask framework compared with a Masked DLM baseline.}
    \label{fig:diffumask}
\end{figure*}
\subsection{DiffuMask}
\label{section:diffumask}
\paragraph{Evaluation and Filtering.}
To construct a high-quality dataset, we filter the pruned prompt candidates to retain only those that yield measurable performance gains. Specifically, a candidate is kept if the final pruned prompt achieves higher accuracy than its corresponding full and fewer-shot prompt on the evaluation split. To further enrich the dataset, we include samples from intermediate states along the pruning trajectories, which capture partial pruning behaviors and provide additional supervision signals. After filtering and aggregation, this process yields a compact yet diverse corpus of full–pruned prompt pairs. We divide this dataset into a training set for finetuning and a validation set for hyperparamters selection. In addition, a separate testing set is constructed from prompts that did not initially show performance improvement. This allows us to evaluate whether \textsc{DiffuMask}-pruned versions of these challenging prompts can surpass their full and fewer-shot counterparts.

\paragraph{Data Representation and Alignment.}
To ensure reproducibility and accurate token-level supervision, we organize the dataset in a structured format that explicitly captures both the full and pruned versions of each prompt. Specifically, each example contains two components: (1) the token sequence for the full 32-shot prompt, tokenized using the \texttt{Llama-3.1-8B-Instruct}~\cite{dubey2024llama} tokenizer, and (2) a binary mask of the same length, where $1$ indicates a retained token and $0$ indicates a removed token.
This representation provides an explicit, position-wise correspondence between the full and pruned prompts. Such alignment enables direct supervision for mask prediction during \textsc{DiffuMask} training and supports precise, token-level comparison between the two prompt variants.

The goal of \textsc{DiffuMask} is to learn an efficient and scalable approximation of the hierarchical pruning behavior captured in the training data. Given a full prompt represented as a token sequence $\mathbf{x} = [x^1, x^2, \dots, x^L]$, the model learns to predict a binary mask sequence $\mathbf{\hat{m}} = [m^1, m^2, \dots, m^L]$ through a diffusion-based iterative denoising process, where  $L$ is the sequence length and $m^i \in \{0,1\}$ indicates whether token $x^i$ is retained.

To align the dataset with the model's tokenization, we convert all prompts from the \texttt{Llama-3.1-8B-Instruct} tokenizer to the \texttt{LLaDA-8B} tokenizer used by the \textsc{DiffuMask} model. Binary retention masks are then realigned to ensure each retained or pruned token precisely matches its counterpart in the new token space, preserving token boundaries and semantics.

The rest of this section describes the overall framework of \textsc{DiffuMask}. As shown in Fig.~\ref{fig:diffumask}, it follows the standard diffusion model structure, comprising a forward process, training objective, and inference procedure. While it shares the same architectural structure as regular masked DLMs, \textsc{DiffuMask} redefines each component to focus on learning token retention behavior instead of reconstructing missing tokens.

\paragraph{Forward Pass.}
DiffuMask operates inversely to masked DLMs. Each training sample begins with a pruned prompt from the dataset, denoted as $\mathbf{x}_{\text{pruned}}=\mathbf{x} \odot \mathbf{m}$, where $\odot$ indicates element-wise product. A timestep variable $t \sim \mathcal{U}(0,1)$ is sampled uniformly to determine the proportion of masked tokens to reveal. The model then constructs a partially revealed sequence $\tilde{\mathbf{x}}_t = \mathbf{x} \odot \mathbf{m}_t$, where $\mathbf{m}_t$ is a stochastic interpolation between the fully pruned mask $\mathbf{m}_0 = \mathbf{m}$ and the fully revealed mask $\mathbf{m}_1 = \{\mathbf{1}\}^L$. As $t$ increases, more redundant tokens are added back into the prompt.

By training on these intermediate states, DiffuMask learns the reverse denoising trajectory that reconstructs the original pruning behavior—i.e., predicting which tokens should be retained given a partially revealed pruned prompt.

\begin{table*}[ht!]
\centering
\setlength{\tabcolsep}{4pt} 
\begin{tabular}{lccccccc}
\toprule
 Methods & Full & Random & Llmlingua & Llmlingua2 & CI & HP & \textbf{DiffuMask} \\
\midrule
Token Counts $\downarrow$ & 5399.1 & 1061.3 & 1153.2 & 1155.2 & 1644.4 & \textbf{660.4} & 1113.8 \\ 
EM Accuracy $\uparrow$ & 76.49 & 60.89 & 67.32 & 70.89 & 76.90 & 75.83 & \textbf{77.07} \\ 
Pruning Inference(min) $\downarrow$ & - & 0 & 0.1 & 0.1 & 0.5 & 1440 & 0.75 \\
\bottomrule
\end{tabular}
\caption{
In-domain evaluation on the GSM8K benchmark under the 
32-shot setting. Reported are average token counts, Exact Match (EM) accuracies and pruning inference of different methods.
}
\label{table:gsm8k}
\end{table*}
\paragraph{Training Objective.}
Masked DLMs are trained to reconstruct masked tokens from corrupted sequences. At each timestep $t$, the model observes a partially masked input $\tilde{\mathbf{x}}_t$ and predicts the original tokens at the masked positions $\mathcal{M}_t$ by minimizing the cross-entropy loss: $\mathcal{L}_{\text{DLM}} = - \sum_{i \in \mathcal{M}_t} \log p_{\theta}(x_i \mid \tilde{\mathbf{x}}_t),$ where $p_{\theta}$ denotes the model’s conditional token distribution. While a masked DLM can in principle generate a \texttt{[MASK]} token, it treats it like any other vocabulary item and learns to replace it with plausible words rather than reason about whether the token itself should be kept or removed. 
As a result, a standard masked DLM cannot directly model token retention behavior, which is inherently a binary decision task.
In contrast, \textsc{DiffuMask} explicitly learns this retention behavior, by predicting a binary mask indicating token retention.
At each timestep $t$, the model takes a partially revealed pruned prompt $\tilde{\mathbf{x}}_t$ from the forward process and outputs a predicted mask $\hat{m}_t$ estimating the probability of retaining each token. The model is trained using binary cross-entropy (BCE) loss, $\mathcal{L}_{\text{BCE}}$, computed between $\hat{m}_t$ and the ground-truth binary mask $m_t$.

To avoid degenerate behavior (i.e., always predicting token removal), $\mathcal{L}_{\text{BCE}}$ is computed over all non-mask tokens—covering both remaining and newly revealed tokens—so the model learns both retention and removal dynamics.

Because pruning labels are imbalanced (removed tokens: retrained tokens $\approx 1:6$), we introduce an \emph{anti-mask penalty} that discourages false removals and improves recall of informative tokens:
\vspace{-0.3cm}
{
\setlength{\belowdisplayskip}{0.3em}
\setlength{\belowdisplayshortskip}{0.3em}
\begin{align*}
\mathcal{L}_{\text{anti-mask}} 
&= - \lambda_{\text{mask}} \cdot 
\frac{1}{|\mathcal{I}|}
\sum_{i \in \mathcal{I}}
\log p_{\text{correct}}(x^i),
\end{align*}
}
where $\mathcal{I}$ denotes incorrectly masked tokens (i.e., $\hat{m}^i_t = 0$ but $m^i_t = 1$), $\lambda_{\text{mask}}$ controls the penalty strength, and $p_{\text{correct}}(x^i)$ is the model’s predicted probability of correctly retaining token $x^i$. 

The final training objective combines both components: $\mathcal{L}_{\text{total}} = \alpha \, \mathcal{L}_{\text{BCE}} + (1 - \alpha) \, \mathcal{L}_{\text{anti-mask}}$, where $\alpha$ balances the reconstruction and penalty terms. This objective enables \textsc{DiffuMask} to jointly model retention and removal decisions while preventing collapse to trivial pruning.
\paragraph{Inference.}
After training, \textsc{DiffuMask} follows the inference procedure of \texttt{LLaDA}, using $64$ diffusion steps. Instead of applying a hard top-1 decision for each token, we adopt a \textit{top-$k$ mask prediction} strategy: a token is pruned if the predicted probability of the \texttt{[MASK]} token ranks within the top-$k$ probabilities in the model’s output distribution. This enables the model to prune tokens more aggressively. To further control pruning intensity, we introduce a \textit{confidence threshold} $\tau$ that specifies the minimum mask probability required to remove a token. Tokens with mask probabilities below $\tau$ are retained, providing a tunable trade-off between compression strength and reasoning fidelity. By adjusting $k$ and $\tau$, users can flexibly balance efficiency and performance at inference time.

\section{Results}
\label{sec:results}
We begin by evaluating the effectiveness of \textsc{DiffuMask} on the testing set described in Section~\ref{section:dataset} (Evaluation and Filtering). This in-domain evaluation measures both the accuracy of pruned prompts and the degree of prompt compression achieved by \textsc{DiffuMask}, compared against established pruning and optimization baselines, including packaged, off-the-shelf methods \textsc{LLMLingua} and \textsc{LLMLingua2}~\cite{jiang2023llmlingua, pan2024llmlingua}, the shot-level pruning method \textsc{CoT-Influx} (~\textsc{CI}), the full hierarchical pruning pipeline (\textsc{HP}, combining \textsc{CoT-Influx} and \textsc{TA-Pruning}), and the original uncompressed 32-shot prompts.

We then assess the generalizability of the finetuned \textsc{DiffuMask} model through two complementary evaluations. The out-of-domain evaluation tests whether \textsc{DiffuMask} (which is trained on GSM8K dataset) generalizes and transfers to completely different datasets and task types beyond mathematical reasoning, while the cross-model evaluation examines whether the pruned prompts generated by \textsc{DiffuMask} (which uses \texttt{Llama-3.1-8B-Instruct} for training set generation) transfer effectively across different LLMs, both within and beyond the LLaMA family. 

All experiments are conducted on A100 GPUs. The finetuning and inference hyperparameters for both in-domain and out-of-domain tasks are provided in Appendix~\ref{appendix:finetune_hparams},\ref{appendix:hyperparameter_search}, respectively.


\begin{table*}[ht!]
\centering
\setlength{\tabcolsep}{10pt}
\begin{tabular}{l|c|c|c||c|c}
\toprule
\multirow{2}{5em}{\textbf{Dataset}} &
\multirow{2}{2.5em}{\textbf{Shots}} &
\multicolumn{2}{c||}{\textbf{Token Counts}} &
\multicolumn{2}{c}{\textbf{Accuracy}} \\
 &  &
\textbf{Full} &
\textbf{DiffuMask} &
\textbf{Full} &
\textbf{DiffuMask} \\
\hline
Yelp5 & 1
& 877.7
& 285.4 (↓67.5\%)
& 61.88
& \textbf{62.93 (+1.05)} \\
      & 5
& 4800.7
& 149.6 (↓96.9\%)
& 50.79
& \textbf{62.91 (+12.12)} \\
\midrule
Yahoo & 1
& 533.3
& 224.8 (↓57.8\%)
& 60.56
& \textbf{63.44 (+2.88)} \\
      & 5
& 2484.4
& 759.8 (↓69.4\%)
& 51.96
& \textbf{63.99 (+12.03)} \\
\midrule
AG's News & 1
& 227.0
& 224.3 (↓1.2\%)
& \textbf{80.28}
& 80.23 (-0.05) \\
          & 5
& 1144.1
& 559.8 (↓51.1\%)
& 61.84
& \textbf{82.50 (+20.66)} \\
          & 20
& 4599.5
& 2140.0 (↓53.5\%)
& 25.37
& \textbf{84.40 (+59.03)} \\
\bottomrule
\end{tabular}
\caption{
Out-of-domain evaluation on classification benchmarks. 
Relative token reduction (↓) and accuracy change ($\pm$) are shown with respect to the full prompt.
}
\label{table:ood}
\end{table*}
\paragraph{In-Domain Evaluation.}
We apply the finetuned \textsc{DiffuMask} model to prune 32-shot prompts from the testing set and then use the resulting pruned prompts as in-context exemplars for the GSM8K test set, which contains 1{,}319 queries. We report Exact Match (EM) accuracy of the predicted results. This setup directly measures whether the pruned prompts generated by \textsc{DiffuMask} can preserve accuracy when applied to unseen problems.

As shown in Table~\ref{table:gsm8k}, \textsc{DiffuMask} achieves accuracy comparable to, and slightly higher than, both the \textsc{Full} and \textsc{CI} baselines, while reducing prompt length by approximately 80\% and 30\%, respectively. Although the accuracy gain over \textsc{HP} is modest (about 1\%), \textsc{DiffuMask} provides orders-of-magnitude acceleration. Whereas \textsc{TA-Pruning} requires \textbf{10-48 hours} to process a single prompt, \textsc{DiffuMask} perform pruning through a multi-step denoising process that masks tokens in parallel, typically completing within about \textbf{one minute}. It also outperforms packaged, off-the-shelf pruning baselines \textsc{LLMLingua} and \textsc{LLMLingua2} by a wide margin in accuracy (77.07 vs.\ 67.32 and 70.89), while achieving comparable compression and similar pruning latency.

Overall, these results indicate that \textsc{DiffuMask} effectively distills the hierarchical pruning dynamics of its teacher system while eliminating its sequential and computational overhead, enabling efficient and scalable prompt compression in practice.

\paragraph{Out-of-Domain Evaluation}
To evaluate whether the learned pruning policy generalizes across domains and task formats, we apply the same \textsc{DiffuMask} model to prune few-shot prompts from several out-of-domain (OOD) classification datasets, including Yelp-5~\cite{asghar2016yelp}, Yahoo Answers~\cite{labrou1999yahoo}, and AG’s News~\cite{zhang2015character}. We construct few-shot prompts from each dataset and prune them using \textsc{DiffuMask}. The pruned prompts are then compared against the original full prompt baseline. All prompts are evaluated on the same backbone model (\texttt{Llama-3.1-8B-Instruct}) under identical setup.

Table~\ref{table:ood} reports results for one-shot and five-shot settings across all datasets and a twenty-shot setting for AG’s News. Across configurations, \textsc{DiffuMask}-pruned prompts consistently outperform the full prompt baseline, achieving higher accuracy despite substantial token reduction. On Yahoo Answers (five-shot), \textsc{DiffuMask} boosts accuracy by \textbf{12.03\%} ($51.96\%$ to $63.99\%$) while shortening prompts by nearly $70\%$. On AG’s News, accuracy rises from $61.84\%$ to $82.50\%$ (\textbf{20.66\%}) in the five-shot setting and from $25.37\%$ to $84.40\%$ (\textbf{59.03\%}) in the twenty-shot setting, even with roughly half the tokens retained.

These results demonstrate that \textsc{DiffuMask} generalizes well beyond its training distribution. Trained solely on mathematical reasoning data, it effectively transfers to various classification tasks via capturing task-agnostic patterns of contextual redundancy rather than dataset-specific heuristics. This strong cross-domain transfer highlights \textsc{DiffuMask} as a scalable and task-agnostic framework for prompt pruning in LLMs.

\paragraph{Cross-Model Evaluation}
To further assess generalization across model architectures, we evaluate \textsc{DiffuMask}-pruned prompts on a range of SOTA LLMs, both within and beyond the Llama family. Specifically, we use GSM8K prompts pruned by the finetuned \textsc{DiffuMask} model and directly apply them—without any model-specific adaptation—to eight distinct LLMs: \texttt{GPT-4o}, \texttt{GPT-4.1}, \texttt{GPT-4.1-nano}, \texttt{Gemini-2.5-Flash}, \texttt{Gemini-2.5-Pro}, \texttt{Grok-3}, \texttt{Llama-3.3-70B}, and \texttt{Llama-4-Maverick}. Each model is evaluated on the GSM8K test set under identical inference setup, and we report reasoning accuracy using full, and \textsc{DiffuMask}-pruned prompts.

As shown in Table~\ref{table:modelanalysis}, \textsc{DiffuMask}-pruned prompts largely maintain accuracy across all model families. The largest gains appear for models architecturally aligned with the source model used during training—for example, \texttt{Llama-3.3-70B} achieves a $+3.4\%$ improvement over the full prompt. For other architectures, the pruned prompts achieve comparable or slightly lower accuracy while still yielding substantial reductions in prompt length. 
\begin{table}[h]
\centering
\setlength{\tabcolsep}{15pt}
\begin{tabular}{lccc}
\toprule
\textbf{Model} & \textbf{Full} & \textbf{DiffuMask} \\
\midrule
GPT4o         &  90.57 & \textbf{92.39}(+1.8) \\ 
GPT4.1         &  90.93 & \textbf{92.76}(+1.8) \\ 
GPT4.1nan      & \textbf{83.95} & 83.48(-0.5) \\ 
Gem2.5Fl  & \textbf{97.03} & 96.40(-0.6) \\
Gem2.5Pro  & 98.12 & \textbf{98.28} (+0.2) \\
GROK 3  & \textbf{97.86} & 96.50(-1.4)\\
Llam3.3-70 & 89.79 &  \textbf{93.23}(+3.4) \\ 
Llam4-Mav & 96.56 &  \textbf{97.03}(+0.5) \\ 
\bottomrule
\end{tabular}
\caption{
\textbf{Cross-model evaluation.} 
Accuracy (\%) on the GSM8K test set using full and 
\textsc{DiffuMask}-pruned prompts across eight LLMs. (GPT4.1nan  = GPT-4.1-nano; Gem2.5Fl = Gemini-2.5-Flash; Gem2.5Pro = Gemini-2.5-Pro; Llam3.3-70 = Llama-3.3-70B-Instruct; Llam4-Mav = Llama-4-Maverick)
}
\label{table:modelanalysis}
\end{table}
These results suggest that \textsc{DiffuMask} captures architecture-independent saliency patterns, enabling the generation of concise yet semantically informative prompts that transfer effectively across diverse LLMs. This cross-model robustness highlights \textsc{DiffuMask} as a practical and model-agnostic pruning mechanism for enhancing efficiency and maintaining reasoning quality without requiring model-specific retraining.

\paragraph{Patterns in Pruned vs. Original Tokens} We analyzed tokens from the original and pruned prompts to confirm that the pruning process is non-trivial. The removed tokens are not limited to specific grammatical categories such as conjunctions or auxiliary verbs. Instead, the overall tag distribution of pruned words closely matches the original across multiple datasets, showing the process does not disproportionately target any word type. Our detailed analysis is provided in Appendix~\ref{sec:tokenanalysis}.
\vspace{-0.2cm}
\section{Conclusion}
\vspace{-0.2cm}
ICL and CoT enable LLMs to perform complex reasoning without parameter updates, but they often lead to excessively long and inefficient prompts. While prompt compression aims to address these inefficiencies, most existing pruning approaches rely on sequential decisions, which are computationally expensive and difficult to scale in practice. 

We present a pipeline of constructing a full–pruned prompt dataset that integrates hierarchical shot- and token-level pruning signals, enabling direct supervision for training efficient pruning models. Building on this, we proposed \textsc{DiffuMask}, a diffusion-based prompt compression framework that learns to predict token retention masks through an iterative denoising process. Unlike conventional pruning methods, \textsc{DiffuMask} efficiently removes multiple redundant tokens at each step, substantially accelerating compression while preserving essential reasoning context.

Empirical results demonstrate that \textsc{DiffuMask} achieves up to an 80\% reduction in prompt length without sacrificing—and in some cases even improving—task accuracy. It consistently performs well across three complementary settings: \emph{in-domain} reasoning on GSM8K, \emph{out-of-domain} generalization to diverse classification tasks, and \emph{cross-model} transfer across different LLM architectures.

Overall, \textsc{DiffuMask} provides a scalable and generalizable framework for balancing reasoning quality and inference efficiency in prompt-based LLMs, offering a promising path toward practical and performant LLM deployment.
\vspace{-0.2cm}
\section*{Limitations}
\vspace{-0.2cm}
While \textsc{DiffuMask} achieves substantial efficiency and generalization gains, some limitations remain. 
First, constructing the full–pruned prompt dataset is computationally expensive: generating each training pair via hierarchical pruning, especially TA-Pruning, can take from 10-48 hours. This high cost limits the scale and diversity of available supervision data and motivates future work on more efficient or self-supervised data generation. Second, the diffusion-based pruning process still involves 64 iterative steps during inference, which, although significantly faster than sequential pruning, is not yet real-time. Third, the model’s performance depends on the pruning behaviors of its teacher methods, which may introduce inductive bias from those heuristics. Finally, our evaluation primarily focuses on reasoning and classification tasks; extending \textsc{DiffuMask} to domains such as dialogue, summarization, and multimodal reasoning remains an open avenue for future exploration.


\bibliography{ref}

\newpage
\appendix

\section{Dataset Construction Details}
\label{appendix:dataset_construction}
This section provides extended details of the dataset construction pipeline described in Section~\ref{section:dataset}, including the procedures for shot-level pruning, token-level pruning, and the subsequent evaluation and filtering stages. Together, these processes produce a high-quality corpus of full–pruned prompt pairs used for fine-tuning \textsc{DiffuMask}.

\subsection{Shot-Level Pruning (Extended Details)}
\label{appendix:shot_pruning}
The shot-level pruning stage aims to reduce the number of in-context exemplars within each prompt while preserving, or even improving, reasoning accuracy. We explore three strategies: a reinforcement learning (RL)–based pruner following \textsc{CoT-Influx}~\cite{huang2024fewer}, and two random selection baselines that provide non-learned alternatives.

\paragraph{RL-Based Shot-Level Pruner.}
Following the setup of \textsc{CoT-Influx}, we train a RL–based pruner to select the most informative exemplars for each query from a pool of candidates.
Training is conducted on the \emph{pruning split} of GSM8K (6{,}000 samples), divided into a \textbf{candidate set} and a \textbf{query set} with a 4:1 ratio—4{,}800 candidates (80\%) and 1{,}200 queries (20\%). In each RL episode, one query is randomly sampled from the query set, and 32 candidate exemplars are retrieved from the candidate set based on semantic similarity computed via the \texttt{SentenceTransformer} module in \texttt{OpenICL}~\cite{wu2023openicl}.
This module encodes both queries and exemplars into a shared embedding space and retrieves the top-32 nearest neighbors according to cosine similarity. The pruner is trained for three epochs, and each episode is repeated 10 times to generate diverse pruned prompts from the learned selection policy.

\paragraph{Alternative Shot-Level Pruning Strategies.}
Although our shot-level pruning is inspired by \textsc{CoT-Influx}~\cite{huang2024fewer}, this stage in our pipeline is modular and can be replaced by other selection strategies. To demonstrate this flexibility and assess whether the RL-based approach offers meaningful improvements over simpler baselines, we construct two additional random-pruning variants:

\begin{itemize}
    \item \textbf{Random Variable-$k$ Selection.}  
    For each 32-shot prompt, we randomly select a variable number of exemplars $k$ such that the \emph{average} retained shot count matches that of the trained pruner (approximately 10). This procedure yields 100 additional prompt pairs for direct comparison.
    \item \textbf{Random Fixed-$k$ Selection.}  
    In this variant, we first randomly generate 32 exemplars for each query from the pruning split. 
    Then, among these 32 candidates, we randomly select $k$ exemplars to form the fewer-shot prompt. 
    Each resulting prompt is evaluated on evaluation split of GSM8K (1273 samples), and the corresponding performance is reported in Table~\ref{tab:random_top32}.  
    According to this result, we select $k = 5$, which achieves the best trade-off between accuracy and prompt length. 
    We thus adopt 5-shot pruning for 600 additional prompts.
\end{itemize}

\begin{table}[t]
\centering
\setlength{\tabcolsep}{10pt} 
\begin{tabular}{c|cc}
\toprule
\# Shots & Token Counts & Accuracy (\%) \\ 
\midrule
1  & 183  & 73.38 \\
2  & 377  & 74.44 \\
5  & 857  & \textbf{75.65} \\
10 & 1704 & 75.64 \\
15 & 2571 & 75.29 \\
20 & 3412 & 75.05 \\
25 & 4284 & 74.78 \\
32 & 5479 & 74.89 \\
\bottomrule
\end{tabular}
\caption{Performance of top-32 random selection with 100 data points.}
\label{tab:random_top32}
\end{table}

For the first two strategies—the RL-based pruner and the random variable-$k$ selection—pruning begins from the same set of \textbf{32-shot full prompts}, ensuring a direct and fair comparison. In contrast, the random fixed-$k$ variant independently samples 32 exemplars for each query before pruning, introducing additional randomness in exemplar composition. Overall, we generate \textbf{100} prompts pruned by the RL-trained pruner, \textbf{100} by random variable-$k$ selection, and \textbf{600} by random fixed-$k$ selection, resulting in a total of \textbf{800 fewer-shot prompts} for the subsequent token-level pruning stage.

\paragraph{Effectiveness of Shot-Level Pruning.}
Table.~\ref{tab:shot_pruning_results} includes the comparison of all three shot-level pruning strategies with their full prompt counterparts. All pruning methods achieve clear performance improvements compared to the original 32-shot baseline, demonstrating that removing redundant exemplars can enhance reasoning accuracy. 

\begin{table}[ht!]
\centering
\setlength{\tabcolsep}{4pt} 

\begin{tabular}{l|cc|cc}
\toprule
Method & \multicolumn{2}{c|}{Full} & \multicolumn{2}{c}{Fewer-shot} \\ 
\midrule
 & Shots & Acc & Shots & Acc \\ 
\midrule
RL Pruner (100) & 32 & 74.42 & 10.3 & 75.32 \\
Rand Var-$k$ (100) & 32 & 74.42 & 10.3 & 75.33 \\
Rand Fix-$k$ (600) & 32 & 74.84 & 5 & 75.72 \\
\bottomrule
\end{tabular}
\caption{Shot-level pruning comparison on GSM8K.}
\label{tab:shot_pruning_results}
\end{table}

However, the RL-based pruner does not offer significant advantages over random selection: both the variable-$k$ and fixed-$k$ random variants perform on par or even slightly better while requiring no additional training. This result suggests that the learned RL policy provides limited additional benefit beyond stochastic exemplar selection. Consequently, the random selection strategies—particularly the fixed-$k$ variant with $k=5$—offer a simpler and more efficient alternative for constructing fewer-shot prompts, streamlining our overall pruned prompt generation pipeline.

\begin{algorithm}[ht!]
\caption{\textsc{TA-Pruning}}
\label{alg:ta_pruning}
\KwIn{Prompt $X = [x^1, \ldots, x^L]$, search split $\mathcal{D}$, performance function $f$, language model $\mathcal{M}$, threshold $\delta$}
\KwOut{Optimized prompt $S$}
\BlankLine
$f^{X} = f(X; \mathcal{D}, \mathcal{M})$ 
\tcp*{Measure baseline performance $f^{X}$ of $X$ on $\mathcal{D}$ using $\mathcal{M}$}
$T \leftarrow X$ \tcp*{Current prompt}
$S \leftarrow X$ \tcp*{Optimal prompt}
$f^{\text{optimal}} \leftarrow f^{X}$\;
\Repeat{Converged}{
    \text{Converged = True}\;
    $K = ||T||$ \tcp*{Number of tokens}
    \For{$i = 1$ \KwTo $K$}{
        $P = T - T^i$ \tcp*{Remove token $T^i$}
        $f^{P} = f(P; \mathcal{D}, \mathcal{M})$ \;
        \uIf{$f^{P} > f^{\text{optimal}}$}{
            $T \leftarrow P$, $S \leftarrow P$\;
            $f^{\text{optimal}} \leftarrow f^{P}$\;
            \text{Converged = False}\;
        }\uElseIf{$f^{P} > f^{\text{optimal}} \times \delta$}{
            $T \leftarrow P$ \tcp*{Within threshold}
            \text{Converged = False} \;
        }
    }
}
\Return{$S$}
\label{appendix:pseudo_code}
\end{algorithm}

\subsection{Token-level Pruning (TA-Pruning)}
\label{appendix:token_pruning}
The token-level pruning stage performs fine-grained compression by selectively removing unnecessary tokens within each exemplar. We adopt the \textsc{TA-Pruning} algorithm introduced in \textsc{PromptQuine}~\cite{wang2025evolving}, which formulates prompt compression as a threshold-accepting optimization process. Given a prompt $X = [x^1, \ldots, x^L]$, the algorithm iteratively removes individual tokens and evaluates the resulting prompt using a performance function $f$. If the pruned prompt achieves higher performance than the current best—or remains within a tolerance threshold $\delta$—the update is accepted. This iterative procedure continues until convergence, yielding a compact prompt that balances accuracy and efficiency. The procedure is summarized in Algorithm~\ref{alg:ta_pruning}.

We use Exact Match (EM) accuracy as the performance function $f$, computed over the \emph{search split} derived from the GSM8K training data (200 samples). The underlying language model $\mathcal{M}$ is \texttt{LLaMA-3.1-8B-Instruct}, and the threshold parameter is set to $\delta = 0.95$. Each prompt is optimized independently until convergence. Further algorithmic details can be found in the original \textsc{PromptQuine} paper~\cite{wang2025evolving}.

\subsection{Evaluation and Filtering (Extended Details)}
\label{appendix:filtering}
The filtering and evaluation stage aims to retain high-quality pruned prompts that demonstrate measurable improvements over their full and fewer-shot counterparts. 
We then use these selected high-quality examples to construct the training and validation sets for \textsc{DiffuMask}, while the unselected examples prompts are reserved as the testing set. 

Specifically, each pruned prompt candidate is evaluated on the evaluation split, and a prompt is retained if its pruned version achieves higher accuracy than both the full prompt and its corresponding fewer-shot version. We then further expand this pool by including intermediate checkpoints along the TA-Pruning trajectories, which capture gradual pruning behaviors and enrich the supervision signal. 
\begin{table}[t]
\centering
\begin{tabular}{l|cc|c}
\toprule
\textbf{Pruning Group} & \textbf{Improved} & \textbf{TA Traj} & \textbf{Total} \\ 
\midrule
RL Pruner      & 55 & 26 & 81 \\
Rand Var-$k$    & 37 & 24 & 61 \\
Rand Fix-$k$ & 61 & - & 61 \\
\midrule
\textbf{Total}         & \textbf{142} & \textbf{61} & \textbf{203} \\
\bottomrule
\end{tabular}
\caption{Summary of token-level pruning results.}
\label{tab:filtering_summary}
\end{table}
Table~\ref{tab:filtering_summary} summarizes the quantitative outcomes of this process across the three shot-level pruning groups. For both the RL-trained and random variable-$k$ groups (each with 100 samples), roughly half of the prompts yield accuracy improvements, contributing 55 and 37 examples respectively. In addition, 26 and 24 supplementary pairs are extracted from their corresponding TA-Pruning trajectories. The random fixed-$k$ group contains 600 samples and produces 61 improved prompts. We did not perform TA-trajectory exploration for this group due to the high computational cost of searching over the entire trajectories. Altogether, the filtering process retains 203 high-quality full–pruned prompt pairs.

Based on these results, we construct the final dataset splits summarized in Table~\ref{tab:dataset_split_summary}. Among the 203 selected high-quality examples, we allocate the majority to the \textbf{training set} for finetuning \textsc{DiffuMask} and reserve a smaller portion for the \textbf{validation set} to guide inference hyperparameter selection $(\text{top-}k, \tau)$, balancing compression strength and task accuracy. For the \textbf{testing set}, we use the 19 non-selected prompts from the RL-Pruner group—those that did not initially show improvement under TA-Pruning. This setup allows us to evaluate whether the \textsc{DiffuMask}-pruned versions of these more challenging prompts can surpass their full and fewer-shot baselines, providing a fair and rigorous test of generalization.

\begin{table}[ht!]
\centering
\begin{tabular}{l|ccc}
\toprule
\textbf{Pruning Group} & \textbf{Train} & \textbf{Validation} & \textbf{Test} \\ 
\midrule
Trained Pruner      & 81 & --  & 19 \\
Random Var-$k$    & 37 & 24 & -- \\
Random Fix-$k$ & 50 & 11 & -- \\
\midrule
\textbf{Total}         & \textbf{168} & \textbf{35} & \textbf{19}\\
\bottomrule
\end{tabular}
\caption{Dataset composition across training, validation, and testing sets.}
\label{tab:dataset_split_summary}
\end{table}

\section{DiffuMask Finetuning Hyperparameters}
\label{appendix:finetune_hparams}
We finetune \textsc{DiffuMask} using the full–pruned prompt pairs described in Appendix~\ref{appendix:filtering}. 
Training is performed with a batch size of one due to the long input sequences (up to 4096 tokens) and GPU memory constraints. 
The learning rate is set to $1\times10^{-4}$ with a linear warmup over the first 100 steps, followed by linear decay. Each model is trained for 20 epochs. 
The \emph{mask penalty} term ($\lambda_{\text{mask}} = 2.0$) controls the regularization strength that discourages over-masking, while the \emph{balance ratio} ($\beta = 0.8$) adjusts the weighting between reconstruction and mask-prediction losses. Together, these parameters stabilize optimization and maintain a consistent trade-off between compression accuracy and mask sparsity across datasets.
\begin{table}[ht!]
\centering
\begin{tabular}{l|c}
\toprule
\textbf{Hyperparameter} & \textbf{Value} \\
\midrule
Batch size & 1 \\
Epochs & 20 \\
Learning rate & $1\times10^{-4}$ \\
Mask penalty coefficient & 2.0 \\
Balance ratio & 0.8 \\
Maximum sequence length & 4096 \\
Warmup steps & 100 \\
\bottomrule
\end{tabular}
\caption{Finetuning Hyperparameters for DiffuMask.}
\label{tab:finetune_hparams}
\end{table}

\section{Hyperparameter Search}
\label{appendix:hyperparameter_search}
\subsection{In-Domain Hyperparameter Search}
\begin{table*}[t]
\centering
\begin{tabular}{l|cccc|cccc}
\toprule
    & \multicolumn{4}{c|}{\textbf{Token Counts}} & \multicolumn{4}{c}{\textbf{Accuracy}} \\
\midrule
    & \multicolumn{8}{c}{\textbf{Threshold}}    \\
\midrule
\textbf{Top-k} & \textbf{0.0001} & \textbf{0.001} & \textbf{0.01} & \textbf{0.1} & \textbf{0.0001} & \textbf{0.001} & \textbf{0.01} & \textbf{0.1} \\
\midrule
\textbf{2} & 150.4 & 1061.1 & 2950.9 & 4219.7 & 75.31 & 80.76 & 80.95 & 79.75 \\
\textbf{3} & 132.1 & 1023.2 & 2943.3 & 4208.3 & 75.35 & 80.73 & 80.63 & 80.11 \\
\textbf{4} & 130.8 & \textbf{978.5}  & 2891.4 & 4205.0 & 73.37 & \textbf{80.93} & 80.91 & 80.22 \\
\bottomrule
\end{tabular}
\caption{Effect of inference hyperparameters on GSM8K performance.}
\label{table:inf_hyperparameters}
\end{table*}
We perform a grid search over inference hyperparameters on the GSM8K validation set (\textbf{35 samples}) to identify effective configurations for the \textsc{DiffuMask} inference process. During inference, \textsc{DiffuMask} prunes a token if the predicted probability of the \texttt{[MASK]} token ranks within the top-$k$ probabilities of the model’s output distribution \emph{and} exceeds a confidence threshold $\tau$. Increasing $k$ makes pruning \emph{more aggressive}, since the \texttt{[MASK]} token is more likely to fall within the top-$k$, thereby reducing token counts. In contrast, increasing $\tau$ makes pruning \emph{more conservative}, since only tokens with higher mask confidence are removed, leading to longer prompts. Together, these two parameters define the upper (via $k$) and lower (via $\tau$) bounds of pruning intensity, allowing fine-grained control of the balance between compression strength and reasoning fidelity.

Table~\ref{table:inf_hyperparameters} summarizes the GSM8K validation results. 
Lower $\tau$ values (e.g., $10^{-4}$) permit pruning with low mask confidence and thus produce stronger compression, while higher $\tau$ values (e.g., $10^{-1}$) retain more tokens. 
Increasing $k$ consistently strengthens pruning and reduces token counts. 
In our validation set, $\text{top-}k=4$ with $\tau=10^{-3}$ attains the best compression–accuracy trade-off, and this configuration is used for GSM8K experiments.
\begin{table*}[ht!]
\centering
\begin{tabular}{lcccc}
\toprule
 & \multicolumn{2}{c}{\textbf{One-shot}} & \multicolumn{2}{c}{\textbf{Five-shot}} \\
\cmidrule(lr){2-3} \cmidrule(lr){4-5}
\textbf{Top-$k$} & \textbf{Accuracy (\%)} & \textbf{Tokens} & \textbf{Accuracy (\%)} & \textbf{Tokens} \\
\midrule
topk10  & 56.88 & 591.0 & 62.78 & 1555.6 \\
topk20  & 62.78 & 476.8 & 62.44 & 453.2 \\
topk50  & 62.93 & 354.4 & 62.78 & 140.7 \\
topk100  & \textbf{63.04} & \textbf{265.6} & \textbf{63.10} & \textbf{130.2} \\
\bottomrule
\end{tabular}
\caption{Hyperparameter search of DiffuMask on the Yelp dataset under one-shot and five-shot settings.}
\label{tab:yelp_hparam}
\end{table*}

\begin{table*}[ht!]
\centering
\begin{tabular}{lcc|cc}
\toprule
 & \multicolumn{2}{c}{\textbf{One-shot}} & \multicolumn{2}{c}{\textbf{Five-shot}} \\
\cmidrule(lr){2-3} \cmidrule(lr){4-5}
\textbf{Top-$k$} & \textbf{Accuracy (\%)} & \textbf{Tokens} & \textbf{Accuracy (\%)} & \textbf{Tokens} \\
\midrule
topk10  & 59.26 & 353.4 & 63.94 & 1355.0 \\
topk20  & 63.29 & 298.9 & 63.91 & 1082.9 \\
topk50  & \textbf{63.53} & \textbf{206.0} & \textbf{64.09} & \textbf{666.4} \\
topk100 & 63.19 & 170.0 & 63.43 & 400.5 \\
topk200 & 62.77 & 127.3 & 62.50 & 178.1 \\
\bottomrule
\end{tabular}
\caption{Hyperparameter search of DiffuMask on the Yahoo dataset under one-shot and five-shot settings.}
\label{tab:yahoo_hparam}
\end{table*}

\begin{table*}[ht!]
\centering
\begin{tabular}{lcc|cc|cc}
\toprule
 & \multicolumn{2}{c}{\textbf{One-shot}} & \multicolumn{2}{c}{\textbf{Five-shot}} & \multicolumn{2}{c}{\textbf{Twenty-shot}} \\
\cmidrule(lr){2-3} \cmidrule(lr){4-5} \cmidrule(lr){6-7}
\textbf{Top-$k$} & \textbf{Accuracy (\%)} & \textbf{Tokens} & \textbf{Accuracy (\%)} & \textbf{Tokens} & \textbf{Accuracy (\%)} & \textbf{Tokens} \\
\midrule
topk2   & 80.48 & 220.9 & 67.58 & 1078.3 & 62.83 & 4287.7 \\
topk5   & \textbf{80.64} & \textbf{219.2} & 73.93 & 991.8 & 79.40 & 3857.6 \\
topk10  & 79.78 & 214.1 & 65.45 & 970.0 & 78.43 & 3613.7 \\
topk20  & 71.22 & 203.3 & 76.58 & 897.2 & 76.38 & 3174.6 \\
topk50  & 76.37 & 187.7 & 79.64 & 715.6 & \textbf{84.75} & \textbf{2121.9} \\
topk100 & 74.25 & 169.1 & \textbf{83.48} & \textbf{544.3} & 84.53 & 1443.3 \\
\bottomrule
\end{tabular}
\caption{Hyperparameter search of DiffuMask on the AG News dataset under one-shot, five-shot, and twenty-shot settings.}
\label{tab:agnews_hparam}
\end{table*}
\subsection{Out-of-Domain Hyperparameter Search}
\label{appendix:ood_hyperparameter_search}
For out-of-domain evaluation, we follow the same hyperparameter tuning protocol as in the GSM8K in-domain experiment. 
For each dataset and few-shot configuration—\textbf{one-shot}, \textbf{five-shot}, and, where applicable, \textbf{twenty-shot}—we generate \textbf{100 samples} and evenly split them into a \textbf{validation} and a \textbf{testing} subset (50 samples each). 
The validation split is used to perform a grid search over $(\text{top-}k, \tau)$ values to identify the most effective pruning configuration for each task, while the testing split is reserved for final evaluation using the selected hyperparameters. This setup ensures that parameter selection and model evaluation remain strictly isolated, providing a fair assessment of \textsc{DiffuMask}'s generalization ability across domains and varying prompt complexities.

Tables~\ref{tab:yelp_hparam}–\ref{tab:agnews_hparam} summarize the hyperparameter search results for three representative out-of-domain datasets: \textbf{Yelp}, \textbf{Yahoo}, and \textbf{AG News}. Across datasets, increasing $\text{top-}k$ generally reduces the average token count, indicating stronger pruning, while accuracy remains stable or even slightly improves until an over-pruning regime is reached. We observe that moderate pruning settings (e.g., $\text{top-}k$ in the range of 50–100) consistently achieve the best balance between compression and accuracy across both one-shot and few-shot settings.

\subsection{Interpretation of Cross-Domain Trends.}
Interestingly, we observe that the optimal $\text{top-}k$ values for out-of-domain datasets are substantially larger than those identified for GSM8K. 
This discrepancy can be attributed to calibration drift and increased uncertainty in the mask prediction scores under distribution shift. 
When evaluated on in-domain data, \textsc{DiffuMask} produces well-calibrated \texttt{[MASK]} probabilities, allowing effective pruning with small $\text{top-}k$ values. 
In contrast, out-of-domain inputs exhibit softer and less decisive mask distributions, as the diffusion-based predictor encounters token patterns and semantic dependencies not seen during training. 
Larger $\text{top-}k$ values thus compensate for this reduced calibration by broadening the set of candidate tokens considered for pruning, effectively restoring pruning coverage and stability. 
Moreover, OOD datasets such as Yelp and AG News tend to have denser lexical semantics than GSM8K, meaning each token contributes more to meaning construction and requires more conservative pruning decisions. 
Together, these factors explain why broader top-$k$ exploration yields more robust performance across domains.


\section{Analysis on pruned tokens}
\label{sec:tokenanalysis}
We perform token analysis comparing the original and pruned prompts to confirm the prompt pruning process is non-trivial: the pruned words are not of certain types of words, e.g., conjunctions, auxiliary verbs, etc. For this analysis, We use universal Part-of-Speech (POS) tags (see Table ~\ref{tab:postag} for the full list) to classify and count the tag distribution over tokens from both the original and the removed part of the prompts over 10 random samples from Yahoo, Yelp5, and GSM8K datasets.

\begin{figure}[!h]
    \centering
    \includegraphics[width=1\linewidth]{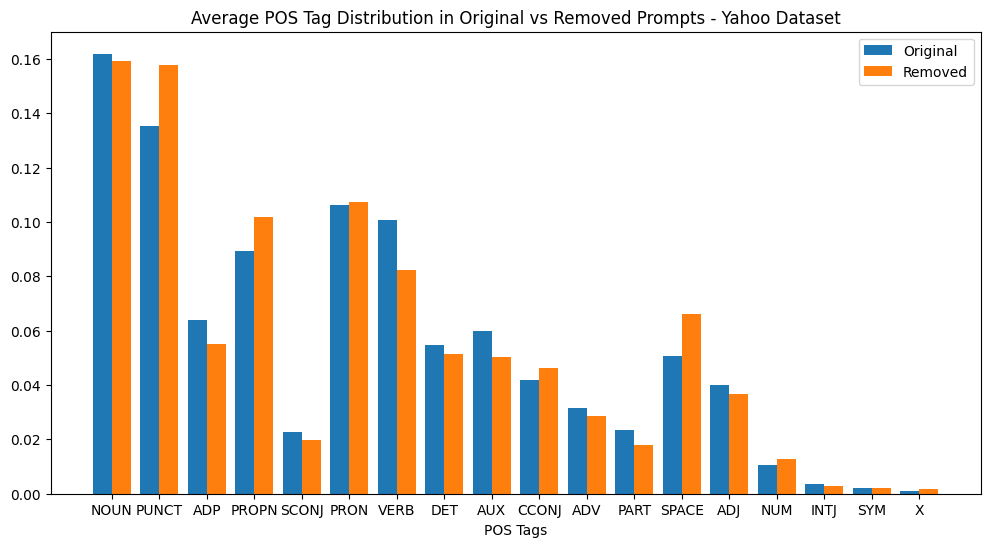}
    \includegraphics[width=1\linewidth]{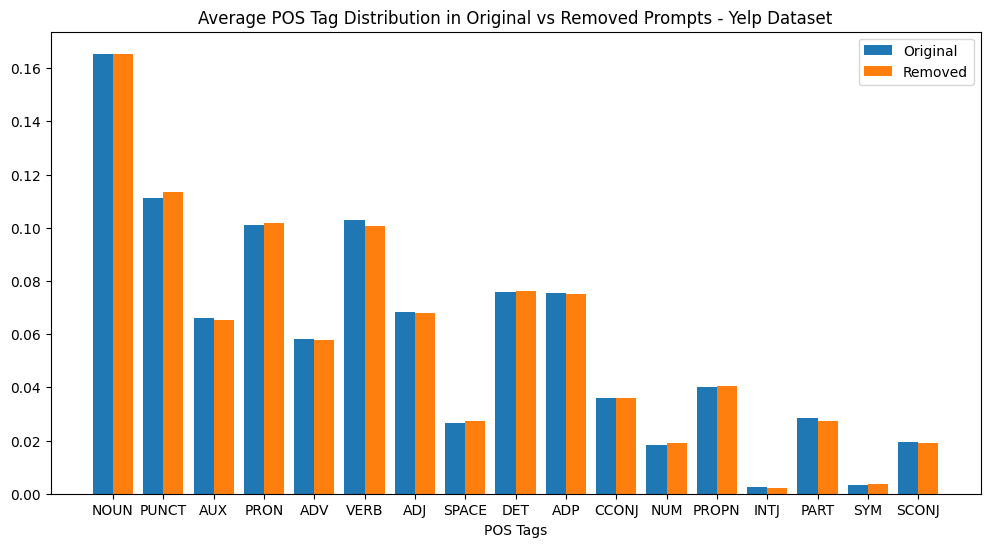}
    \includegraphics[width=1\linewidth]{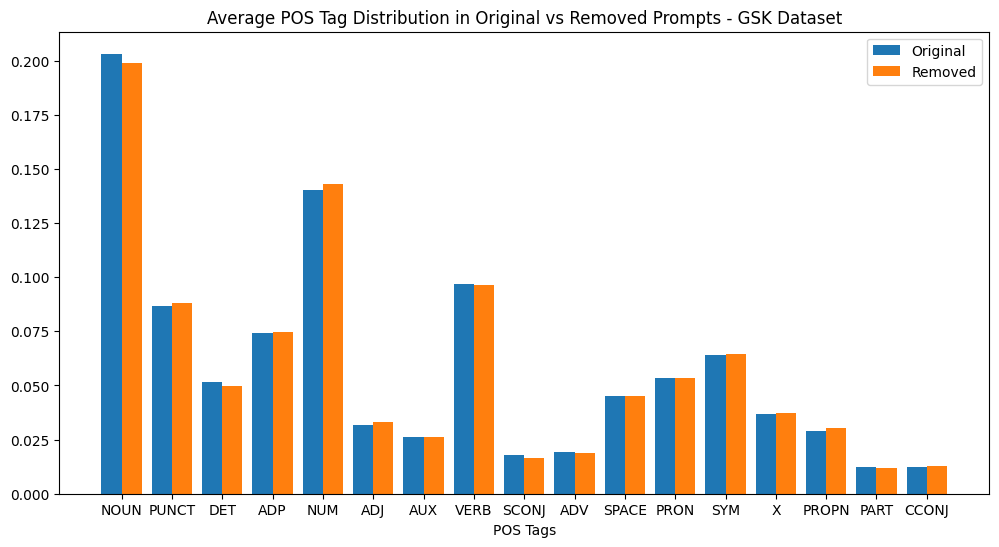}
    \caption{Comparison of POS Tag Distribution (Original Prompts vs Removed Parts of Prompts)}
    \label{fig:tagcomparison}
\end{figure}

The comparisons are shown in Figure ~\ref{fig:tagcomparison}. We can see there are minor deviations for several tags in Yahoo dataset, e.g., punctuations are more frequently pruned than their original distribution, and verbs are more likely retained. However, the overall tag distribution of the parts of the prompts being pruned follows that of the original prompts fairly closely across all 3 datasets, indicating our pruning process is not over-indexing on any type of words and hence not trivial.

\begin{table}[ht!]
\centering
\small
\begin{tabular}{cc}\\
ADJ: & adjective (e.g., big, old)\\
ADP: & adposition (e.g., in, to, during)\\
ADV: & adverb (e.g., very, tomorrow)\\
AUX: & auxiliary verb (e.g., is, has)\\
CCONJ: & coordinating conjunction (e.g., and, but)\\
DET: & determiner (e.g., the, a, some)\\
INTJ: & interjection (e.g., oh, wow)\\
NOUN: & noun (e.g., cat, house)\\
NUM: & numeral (e.g., one, two, first)\\
PART: & particle (e.g., not, up)\\
PRON: & pronoun (e.g., I, he, she)\\
PROPN: & proper noun (e.g., John, London)\\
PUNCT: & punctuation (e.g., ., ,, !)\\
SCONJ: & subordinating conjunction (e.g., if, while)\\
SYM: & symbol (e.g., \$, \%, \&)\\
VERB: & verb (e.g., run, eat, sleep)\\
X: & other (e.g., foreign words, unknown)\\
\end{tabular}
\caption{Universal Part-of-Speech (POS) tags}
\label{tab:postag}
\end{table}

\section{Evaluation metric used for Out-of-Domain datasets}

\begin{table}[!ht]
\centering
\small
\begin{tabular}{p{1.0cm} p{1.9cm} p{3.0cm}}
\textbf{Dataset} & \textbf{Primary Accuracy Measure} & \textbf{What the Accuracy is Checking} \\
\hline
\textbf{AGNews} & \textbf{Top-1 Accuracy} & Whether the model correctly identified the \textbf{single news topic} (World, Sports, Business, or Sci/Tech). \\
\textbf{Yelp5} & \textbf{Star Rating Accuracy} & Whether the model predicted the \textbf{exact star rating} (1, 2, 3, 4, or 5) assigned by the user in their review. \\
\textbf{Yahoo} & \textbf{Topic Classification Accuracy} & Whether the model correctly assigned the text to one of \textbf{10 main categories} (e.g., Health, Sports, Politics). \\
\hline
\end{tabular}
\caption{Accuracy-based evaluation used for out-of-domain datasets.}
\label{tab:ood_accuracy_eval}
\end{table}


To align with evaluation practices commonly used in prior work on these benchmarks, we report classification accuracy on AGNews (news classification), Yelp5 (rating classification), and Yahoo Answers (question/topic classification). These datasets are standard multi-class text classification tasks for which accuracy is the predominant metric in the literature, and their labels are typically assigned using rule-based criteria rather than probabilistic methods. Table~\ref{tab:ood_accuracy_eval} shows the summary for these three datasets.

\end{document}